\documentclass[conference]{IEEEtran}

\usepackage{graphicx}
\usepackage{url}
\usepackage{booktabs}
\usepackage{cite}
\usepackage{amsmath}

\title{When Does a Classifier Help an LLM? Classifier-Guided Prompting and Hybrid Classifier--LLM Models for Credit-Default Prediction}

\author{
\IEEEauthorblockN{Rishi Datta}
\IEEEauthorblockA{Amador Valley High School\\ Pleasanton, California, USA\\ rishidatta20@gmail.com}
\and
\IEEEauthorblockN{Lavanya Prahallad}
\IEEEauthorblockA{Research Spark Hub Inc.\\ Dublin, California, USA\\ lavanya@researchsparkhub.com}
}

\begin{document}
\maketitle

\begin{abstract}
Credit-default prediction is an important task in financial decision making. Traditional methods use fitted classifiers such as logistic regression and random forests on tabular features. Large language models (LLMs) have recently been applied to this task through prompting. In this work we study how a fitted classifier and an LLM can be combined for credit-default prediction. We distinguish telling the LLM to imitate a classifier from using the classifier to build the prompt. We hypothesize that a fitted classifier can supply the ranking ability that an LLM prompt lacks. We experiment on the Default of Credit Card Clients dataset, and report recall, F1, and the area under the ROC and precision-recall curves, with bootstrap confidence intervals. We observe that a few-shot LLM has the highest recall (0.47) and F1 (0.50) of any single model but ranks worse than a random forest (AUC-ROC 0.72 against 0.79). Instructing the LLM to imitate a classifier gives no significant change. Pruning the prompt to the classifier's eight most important features raises recall by 0.071 and F1 by 0.032. Adding the classifier's predicted probability to the prompt raises the LLM's AUC-ROC from 0.72 to 0.78, matching the random forest, while keeping 0.118 higher recall than it. The reverse composition, and the use of several classifiers, do not help. We thus recommend a simple classifier-guided prompt for LLM-based credit prediction.
\end{abstract}


\section{Introduction}

Credit-default prediction is an important task in financial decision making. Given a client's repayment history, credit limit, bill amounts, payment amounts, and demographic details, a model estimates the probability that the client will default. Traditional methods treat this as a supervised classification problem. Logistic regression and tree ensembles such as random forests are common and strong baselines, because they capture the nonlinear interactions found in credit data~\cite{yeh2009comparative,breiman2001random,lessmann2015benchmarking}.

Large language models (LLMs) have recently been used for financial decision support, because they follow instructions and produce natural-language explanations~\cite{brown2020language}. Several studies apply LLMs to credit and financial tasks. Feng et al.~\cite{feng2023calm} instruction-tune a credit LLM and compare it with GPT-4. Sanz-Guerrero and Arroyo~\cite{sanzguerrero2025p2p} build a risk indicator from peer-to-peer loan descriptions. Domain models and surveys map the wider area~\cite{yang2023fingpt,li2023finance}, and a review of interpretable credit LLMs organizes architectures and explanation methods~\cite{golec2025review}. Explainability-focused pipelines usually keep the prediction in a conventional model and use the model output for explanation~\cite{nallakaruppan2024credit,bussmann2021explainable}. Most of this work uses text inputs such as loan descriptions, or fine-tuned domain models. We instead study prompting over structured tabular records with a general model.

A separate line of work studies LLMs on tabular prediction directly. TabLLM~\cite{hegselmann2023tabllm} shows that few-shot classification of serialized tables is competitive with trees at very low shot counts. A survey~\cite{fang2024survey} and newer methods extend this through fine-tuning~\cite{rabbani2025transfer} and retrieval-augmented in-context learning~\cite{wen2025scalable}. These studies show that LLMs can predict on tabular data, but they also find tree ensembles stronger as the data grow.

An LLM given tabular records is not the same as a fitted classifier. A fitted classifier learns parameters from labeled data. An LLM can only be instructed at inference time. This raises a practical question that, to our knowledge, has not been isolated for credit risk: when, and how, does a fitted classifier help an LLM, and does the LLM help the classifier? We consider three ways a classifier can be connected to an LLM prompt. The first is imitation: we tell the LLM to reason like the classifier, using a role instruction or feature-importance hints. The second is prompt construction: we use the classifier to decide what goes in the prompt, that is, which features to show and which few-shot examples to use. The third is a hybrid: we feed the classifier's output into the prompt, or the LLM's output into the classifier. Related methods use LLMs to select or engineer features~\cite{jeong2025llmselect,li2024featsel,han2024featllm} and compare trees with LLMs~\cite{huertas2024gbt}. Our prompt-construction condition inverts feature selection: we use a fitted classifier's importances to prune the prompt. Our hybrid conditions test both directions of composition.

We hypothesize that a fitted classifier can supply information that an LLM prompt lacks, in particular the ability to rank clients by risk. We experiment on the Default of Credit Card Clients dataset. We use a fixed set of 2000 held-out clients, and report recall, precision, F1, and AUC, with bootstrap confidence intervals. We compare a few-shot LLM with fitted logistic regression, random forest, and AdaBoost. We then test imitation prompts, classifier-guided feature pruning, few-shot example selection, and hybrid models in both directions. We also study whether diffusion-based augmentation of the minority class helps a fitted classifier, and compare it with SMOTE~\cite{chawla2002smote} and class-weighting; diffusion models such as TabDDPM~\cite{kotelnikov2023tabddpm} and LLM-based synthesizers~\cite{kim2024epic} motivate this comparison. Throughout, we report null and negative results together with the positive ones.

\section{Database used in this study}
We use the \textit{Default of Credit Card Clients} dataset~\cite{yeh2009comparative}: 30{,}000 clients, 23 features (credit limit, sex, education, marital status, age; repayment-status codes \texttt{PAY\_0}--\texttt{PAY\_6}; bill amounts \texttt{BILL\_AMT1--6}; payment amounts \texttt{PAY\_AMT1--6}), and a binary default label. The data are imbalanced: $22.1\%$ default.

We split 80/20 with a fixed seed, \emph{stratified} by label (24{,}000 train / 6{,}000 test). Fitted models train on the full 24{,}000. Because LLM inference is comparatively expensive, all LLM conditions are evaluated on a fixed, stratified subset of 2{,}000 held-out test clients (default rate $22.1\%$); every LLM condition sees the same 2{,}000 clients. The diffusion-augmentation study (Section~\ref{sec:diffusion}), which involves only fitted models, is evaluated on the full 6{,}000-client test set.

\section{Prediction models used in this study}

\subsection{Fitted baselines}
\textbf{Majority} predicts the majority (no-default) class. \textbf{Logistic Regression} is trained on standardized features. \textbf{Random Forest} (RF) uses 100 trees, maximum depth 12. \textbf{AdaBoost} uses 200 estimators. All use a fixed seed.

\subsection{LLM prediction}
The LLM is \texttt{claude-sonnet-5} with extended thinking disabled for near-determinism. Each client's features are serialized as text; the model is given $k{=}12$ balanced labeled few-shot examples (6 default, 6 non-default) drawn from the training set and asked to return, for each client, a JSON object with a binary \texttt{label} and a \texttt{confidence} $\in[0,1]$ interpreted as $P(\text{default})$ (used for AUC). Requests are batched (20 clients per call) and cached. We evaluate the following LLM conditions.

\textbf{Imitation prompting:}
\begin{itemize}
    \item \textbf{L1 (direct):} predict from features and the few-shot examples, no classifier guidance.
    \item \textbf{L2 (full):} ``act as a trained Random Forest'' \emph{and} a list of the RF's top features to weigh.
    \item \textbf{L2a (role only):} ``act as a trained Random Forest,'' no feature list.
    \item \textbf{L2b (features only):} weigh the RF's important features, no role.
\end{itemize}

\textbf{Classifier-guided prompt construction:} (instruction text held constant at the plain L1 wording; only prompt \emph{content} changes).
\begin{itemize}
    \item \textbf{E1 (top-$K$ features):} include only the RF's top-8 features; drop the other 15.
    \item \textbf{E2 (boundary shots):} choose the 12 few-shot examples nearest the RF decision boundary (out-of-bag probability closest to $0.5$), balanced by class.
    \item \textbf{E3:} E1 $+$ boundary shots.
    \item \textbf{E4 (representative shots):} choose the most confidently-correct examples (out-of-bag probability near 1 for defaulters, near 0 for non-defaulters), balanced.
    \item \textbf{E5:} E1 $+$ representative shots.
\end{itemize}
Few-shot verdicts used for selection come from out-of-bag/cross-validated probabilities so they are not overfit.

\textbf{Hybrid classifier-LLM composition:}
\begin{itemize}
    \item \textbf{E6 (RF $\rightarrow$ LLM):} E1 $+$ the RF's predicted probability and 0/1 decision as extra inputs for each client (and each few-shot example, using out-of-bag verdicts).
    \item \textbf{E7 (ensemble $\rightarrow$ LLM):} E1 $+$ the probabilities and decisions of \emph{three} classifiers (RF, logistic regression, AdaBoost).
    \item \textbf{Inverse (LLM $\rightarrow$ RF):} add the LLM's E1 label and confidence as features to the RF; evaluated by 5-fold cross-validation on the 2{,}000 (comparing RF with vs.\ without the LLM feature on identical folds).
\end{itemize}

\subsection{Metrics and uncertainty}
We report accuracy. Because the data are imbalanced, we also report precision, recall, and F1 on the default class, plus AUC-ROC and AUC-PR. We list overall accuracy next to the majority-baseline accuracy (0.779) so it can be read in context. 
We call a difference significant when its 95\% CI excludes zero. The diffusion study reports mean $\pm$ 95\% CI across 5 seeds.

\section{Results}

\begin{figure}[!t]
\centering
\includegraphics[width=0.92\columnwidth]{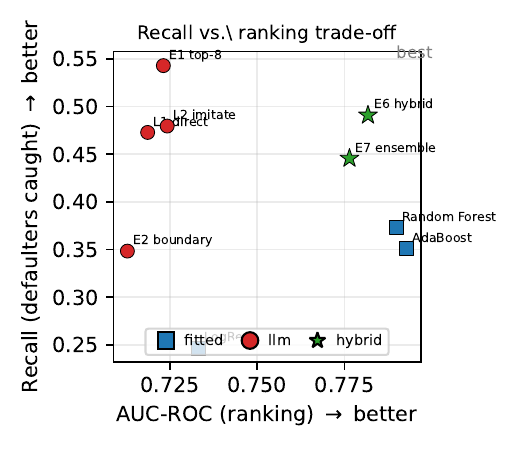}
\caption{Recall vs.\ AUC-ROC across models (N=2000). Fitted models (squares) rank well but catch few defaulters; plain/prompt-engineered LLM conditions (circles) catch more defaulters but rank worse; the hybrid E6 (star) attains Random-Forest-level ranking \emph{and} high recall.}
\label{fig:tradeoff}
\end{figure}

Figure~\ref{fig:tradeoff} shows the trade-off between recall and ranking. An observation of the figure shows that the fitted models lie in the high-AUC, low-recall region, the LLM conditions lie in the low-AUC, high-recall region, and the hybrid E6 lies in the high-AUC, high-recall region.

\subsection{LLM vs.\ fitted models, and imitation prompting}
Table~\ref{tab:main} reports all fitted baselines and the imitation-prompting LLM conditions on the 2{,}000-client set.

\begin{table}[!t]
\caption{Fitted baselines and imitation-prompting LLM conditions (N=2000, default rate 22.1\%). Best value per column in bold.}
\label{tab:main}
\centering
\setlength{\tabcolsep}{3pt}
\footnotesize
\begin{tabular}{lcccccc}
\toprule
Model & Acc & Prec & Rec & F1 & AUROC & AUPRC \\
\midrule
Majority          & 0.779 & --    & 0.000 & 0.000 & --    & --    \\
Logistic Reg.     & 0.811 & \textbf{0.708} & 0.247 & 0.366 & 0.733 & 0.512 \\
Random Forest     & 0.821 & 0.668 & 0.373 & 0.479 & 0.790 & \textbf{0.577} \\
AdaBoost          & \textbf{0.823} & 0.698 & 0.351 & 0.467 & \textbf{0.793} & 0.571 \\
\midrule
L1: direct LLM    & 0.789 & 0.525 & 0.473 & 0.498 & 0.719 & 0.481 \\
L2: role+features & 0.790 & 0.526 & 0.480 & 0.502 & 0.724 & 0.478 \\
L2a: role only    & 0.795 & 0.540 & 0.477 & 0.507 & 0.722 & 0.487 \\
L2b: features only& 0.790 & 0.526 & 0.489 & 0.506 & 0.728 & 0.486 \\
E1: top-8 feats   & 0.786 & 0.515 & \textbf{0.543} & \textbf{0.529} & 0.723 & 0.470 \\
\bottomrule
\end{tabular}
\end{table}

We make two observations. First, the LLM trades precision for recall. The direct LLM (L1) flags about 20\% of clients as defaulters, against the RF's 12\%. It gives the highest recall (0.47) and, with E1, the highest F1 (0.53) of any model, above the fitted RF (0.48). The fitted models keep higher precision, accuracy, and AUC. So the LLM is strong on recall but weaker on ranking. Second, imitation prompting does not help. The paired differences of L2, L2a, and L2b against L1 are not significant on F1, AUC-ROC, or recall (all 95\% CIs include zero; for example, L2$-$L1 F1 $=+0.004$, CI $[-0.013,+0.020]$). Neither the ``act as a Random Forest'' role nor an explicit feature-importance list changes performance.

\subsection{Classifier-guided prompt construction}
Table~\ref{tab:paired} reports paired differences versus L1 for the prompt-construction and hybrid conditions.

\begin{table*}[!t]
\caption{Prompt-construction and hybrid conditions vs.\ the L1 baseline (N=2000). Each cell gives the absolute metric and, in parentheses, the paired difference from L1 with its 95\% bootstrap CI (1000 resamples). \textbf{Bold} marks a significant difference (CI excludes 0). The L1 row gives the absolute reference values.}
\label{tab:paired}
\centering
\footnotesize
\setlength{\tabcolsep}{5pt}
\resizebox{\textwidth}{!}{%
\begin{tabular}{lccc}
\toprule
Condition & F1 ($\Delta$ vs.\ L1 [95\% CI]) & AUC-ROC ($\Delta$ vs.\ L1 [95\% CI]) & Recall ($\Delta$ vs.\ L1 [95\% CI]) \\
\midrule
L1 (baseline)      & 0.498 & 0.719 & 0.473 \\
E1 top-8 feats     & 0.529 (\textbf{+.032 [.006,.056]}) & 0.723 (+.005 [$-$.013,.022]) & 0.543 (\textbf{+.071 [.040,.102]}) \\
E2 boundary shots  & 0.441 (\textbf{$-$.057 [$-$.088,$-$.025]}) & 0.713 ($-$.006 [$-$.018,.006]) & 0.348 (\textbf{$-$.125 [$-$.158,$-$.092]}) \\
E3 top-8+boundary  & 0.487 ($-$.011 [$-$.039,.015]) & 0.725 (+.006 [$-$.013,.023]) & 0.428 (\textbf{$-$.045 [$-$.075,$-$.015]}) \\
E4 repr.\ shots    & 0.469 (\textbf{$-$.029 [$-$.054,$-$.006]}) & 0.720 (+.001 [$-$.009,.013]) & 0.403 (\textbf{$-$.070 [$-$.097,$-$.044]}) \\
E5 top-8+repr.     & 0.517 (+.020 [$-$.005,.046]) & 0.730 (+.011 [$-$.004,.027]) & 0.489 (+.016 [$-$.012,.047]) \\
\midrule
E6 (E1+RF)         & 0.524 (+.026 [$-$.000,.054]) & 0.782 (\textbf{+.063 [.039,.085]}) & 0.491 (+.019 [$-$.010,.048]) \\
E7 (E1+RF+LR+ADA)  & 0.510 (+.026$^{\dagger}$) & 0.776 (+.058$^{\dagger}$) & 0.446 ($-$.010$^{\dagger}$) \\
\bottomrule
\end{tabular}}
\\[2pt]
\raggedright \footnotesize $^{\dagger}$E7 differences vs.\ L1 are shown for reference; the text compares E7 against E6.
\end{table*}

Restricting the prompt to the RF's top-8 features (E1) improves recall by $0.071$ and F1 by $0.032$, both significant. E1 has the best F1 of any model in the study (0.529), above the fitted RF. Both example-curation strategies lower recall and F1. Boundary (hardest-case) selection (E2) is the worst ($\Delta$F1 $-0.057$, $\Delta$recall $-0.125$), and representative (prototypical) selection (E4) also hurts ($\Delta$F1 $-0.029$). Combining feature pruning with curated examples (E3, E5) removes E1's gain. In short, a classifier helps by choosing which features the LLM sees, but random balanced few-shot examples beat RF-curated ones.

\begin{figure*}[!t]
\centering
\includegraphics[width=0.86\textwidth]{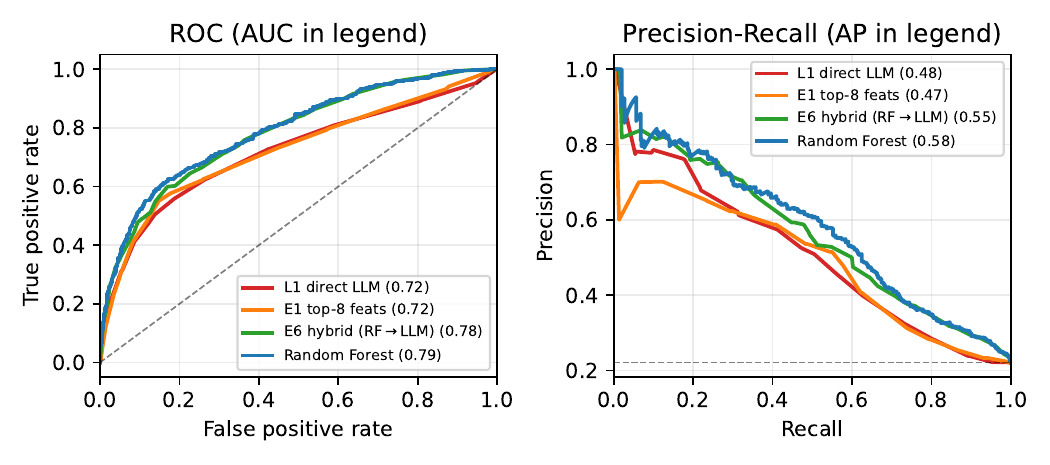}
\caption{ROC (left) and precision-recall (right) curves on the 2000-client test set for the direct LLM (L1), feature-pruned LLM (E1), the hybrid (E6), and the fitted Random Forest. Feeding the RF's probability into the prompt (E6) lifts the LLM's ROC/PR curves to essentially the Random Forest's level.}
\label{fig:roc}
\end{figure*}

\subsection{Hybrid classifier--LLM composition}
\textit{RF $\rightarrow$ LLM (E6) is the best operating point.} Adding the RF's probability and decision to the prompt raises AUC-ROC from 0.719 (L1) to 0.782, a significant gain of $0.063$ that reaches the fitted RF's level (0.790). AUC-PR rises from 0.481 to 0.549. E6 also keeps high recall. Against the fitted RF, E6 improves recall by $0.118$ and F1 by $0.045$ (both significant), at a small AUC cost ($-0.008$). E6 therefore ranks like the Random Forest but catches more defaulters (Table~\ref{tab:hybrid}).

\begin{table}[!t]
\caption{Hybrid and reference models (N=2000). E6/E7 vs.\ RF paired: E6 recall $+0.118$ (sig), F1 $+0.045$ (sig), AUROC $-0.008$ (sig). E7 vs.\ E6: recall $-0.045$ (sig), F1/AUROC n.s.}
\label{tab:hybrid}
\centering
\setlength{\tabcolsep}{3pt}
\footnotesize
\begin{tabular}{lcccccc}
\toprule
Model & Acc & Prec & Rec & F1 & AUROC & AUPRC \\
\midrule
E1 (feats only)        & 0.786 & 0.515 & \textbf{0.543} & \textbf{0.529} & 0.723 & 0.470 \\
E6 (E1+RF)             & 0.803 & 0.561 & 0.491 & 0.524 & 0.782 & 0.549 \\
E7 (E1+RF+LR+ADA)      & 0.811 & 0.595 & 0.446 & 0.510 & 0.776 & 0.536 \\
Random Forest          & 0.821 & 0.668 & 0.373 & 0.479 & \textbf{0.790} & \textbf{0.577} \\
\bottomrule
\end{tabular}
\end{table}

\textit{More classifiers do not help.} Feeding three classifier opinions (RF, logistic regression, AdaBoost; E7) does not improve ranking over the RF-only hybrid (E7 vs.\ E6: $\Delta$AUC-ROC $-0.005$, n.s.; $\Delta$F1 $-0.014$, n.s.) and significantly \emph{reduces} recall ($-0.045$). The three classifiers are highly correlated (all AUC-ROC $\approx 0.79$, all high-precision/low-recall), so their combined signal is redundant and pushes the LLM toward a more conservative consensus.

\textit{The reverse composition does not help.} We add the LLM's E1 output (label and confidence) as features to the RF and evaluate by 5-fold cross-validation on the 2{,}000 clients. This gives no significant change. On identical folds, RF-with-LLM versus RF-without gives $\Delta$recall $+0.002$, $\Delta$F1 $-0.004$, and $\Delta$AUC-ROC $-0.002$ (all not significant). The RF ranks the LLM's confidence as its most important input, yet the metrics do not move. The LLM's signal comes from the same features, so it repeats what the RF already extracts. The useful information flows one way. The classifier gives the LLM ranking ability that the LLM lacks, but the LLM gives the classifier nothing new.

\subsection{Diffusion-based minority augmentation}
\label{sec:diffusion}
Because recall is limited by class imbalance, we test whether augmenting the training data with \emph{synthetic defaulters} from a diffusion model helps a fitted RF, compared with standard remedies. We train a Gaussian denoising diffusion model (with a quantile-normal transform of continuous features) on the defaulter rows and generate synthetic defaulters to rebalance the training set; a detection classifier distinguishes real from synthetic defaulters with AUC $0.80$ (imperfect but usable). Table~\ref{tab:diffusion} reports RF performance on the full 6{,}000-client test set, averaged over 5 seeds.

\begin{table}[!t]
\caption{Random Forest under imbalance remedies (full test, N=6000; mean $\pm$ 95\% CI over 5 seeds).}
\label{tab:diffusion}
\centering
\setlength{\tabcolsep}{3pt}
\footnotesize
\begin{tabular}{lccc}
\toprule
Strategy & Recall & F1 & AUPRC \\
\midrule
original            & 0.351\,$\pm$.004 & 0.458\,$\pm$.003 & 0.556\,$\pm$.003 \\
class\_weight       & \textbf{0.575\,$\pm$.004} & \textbf{0.538\,$\pm$.003} & 0.550\,$\pm$.003 \\
oversample          & 0.561\,$\pm$.005 & 0.537\,$\pm$.003 & 0.547\,$\pm$.001 \\
SMOTE               & 0.541\,$\pm$.006 & 0.511\,$\pm$.005 & 0.518\,$\pm$.004 \\
diffusion           & 0.541\,$\pm$.045 & 0.526\,$\pm$.017 & 0.540\,$\pm$.015 \\
\bottomrule
\end{tabular}
\end{table}

Paired across seeds, diffusion augmentation beats SMOTE, the standard synthetic-minority baseline, on AUC-PR ($+0.022$), precision ($+0.028$), and accuracy ($+0.013$), and ties it on recall and F1. It does not beat simple class-weighting, which is free and gives the highest recall and F1. Diffusion also varies more from run to run. So diffusion is a better synthetic generator than SMOTE for this task, but it is not worth the extra complexity over class-weighting.

\section{Analysis of results}
Our results give a clear picture of how a fitted classifier and an LLM should and should not be combined for imbalanced tabular credit prediction.

Telling an LLM to imitate a classifier does not work. Using the classifier to shape the prompt works, but only in one way: choosing which features to show. Dropping the 15 low-importance columns removes distraction, and this was the only prompt-only change that helped. Curating the few-shot examples, by either difficulty or prototypicality, hurt. Random balanced examples were better. This suggests that example diversity matters more than example ``informativeness'' for this task.

The most useful result is the hybrid. Feeding one strong classifier's probability into the prompt gives the LLM the ranking ability it otherwise lacks (AUC-ROC $0.72\to0.78$) while keeping its higher recall. The result ranks like the Random Forest but catches more defaulters. Random Forest inference costs almost nothing, so this hybrid is cheap to run. The effect is one-directional. The reverse composition (LLM into classifier) and richer ensembles (three classifiers into the LLM) added nothing. A single strong classifier already gives the LLM the signal it needs.

\section{Limitations}
Results use a single LLM (\texttt{claude-sonnet-5}), a single dataset, one 2{,}000-client evaluation draw, and one few-shot draw per condition. Bootstrap CIs capture test-set sampling uncertainty but not LLM-sampling or few-shot-draw variance; the ablation and hybrid conclusions would be strengthened by multiple prompt draws and additional models. The inverse-composition test trains the cross-validated RF on $\sim$1{,}600 rows per fold, below the 24{,}000-row RF, though the RF-with vs.\ RF-without comparison is controlled. The LLM used only the same features available to the classifiers; an LLM with access to information the classifier lacks (e.g., unstructured text) might contribute complementary signal, which we did not test. The diffusion generator treats ordinal repayment codes as continuous; a fully mixed-type generator might improve synthetic quality.

\section{Conclusions}
In this paper we study how to combine a fitted classifier and an LLM for credit-default prediction on the Default of Credit Card Clients dataset. We show that a few-shot LLM has high recall but weak ranking, while a fitted random forest has strong ranking but low recall. We demonstrate that telling the LLM to imitate a classifier does not improve prediction, but that using the classifier to prune the prompt to its important features does. We also show that selecting few-shot examples by decision-boundary proximity or by prototypicality lowers performance, and that random balanced examples are better.

We further demonstrate that a simple hybrid works best. When we add the random forest's predicted probability to the prompt, the LLM ranks as well as the random forest and still catches more defaulters. The advantage of this hybrid is that random forest inference is almost free, so the added cost is small. We show that the reverse composition, and the use of three classifiers instead of one, do not help. Separately, we find that diffusion-based augmentation of the minority class is better than SMOTE but does not beat simple class-weighting.

In summary, for credit-default prediction we recommend pruning the prompt to a classifier's important features and adding one strong classifier's probability to the prompt. In the future we wish to test these findings on more datasets, with more than one language model, and with several few-shot draws, so that the confidence intervals also cover prompt and model variation.

\bibliographystyle{IEEEtran}
\bibliography{references}

\appendices
\section{Prompts, Model Details, and Reproducibility}

\textbf{Model and inference.} All LLM conditions use \texttt{claude-sonnet-5} (Anthropic) with \emph{extended thinking disabled} for near-determinism. Requests set \texttt{max\_tokens} to 2000--3000; no temperature parameter is passed (SDK default). Clients are sent in batches of 20 per request, and every raw response is cached to disk so re-runs reproduce the exact numbers without new API calls. Each response is required to be a strict JSON array and is parsed directly.

\textbf{Few-shot examples.} Every prompt includes $k{=}12$ balanced labeled examples (6 default, 6 non-default) drawn from the training split with seed 42. The same random set is used for all conditions except the exemplar-selection conditions (E2/E4), which instead select 12 examples by out-of-bag RF probability (nearest 0.5 for boundary; nearest 0/1 for representative).

\textbf{Serialization.} Each client is rendered as comma-separated \texttt{name=value} fields. A few-shot line is \texttt{- <fields> -> <label>}; a query line is \texttt{id <k>: <fields>}. Conditions E1/E3/E5 restrict \texttt{<fields>} to the RF's top-8 features (\texttt{PAY\_0, PAY\_2, BILL\_AMT1, PAY\_4, PAY\_3, LIMIT\_BAL, PAY\_AMT1, BILL\_AMT2}). Hybrid conditions append classifier outputs to each line (see Fig.~\ref{fig:instr}).

\textbf{Fitted models and evaluation.} Random Forest (100 trees, depth 12), logistic regression (standardized), AdaBoost (200 estimators); stratified 80/20 split, seed 42; LLM evaluation on a stratified 2{,}000-client subset of the test set. Uncertainty: 1{,}000 bootstrap resamples, paired for between-condition comparisons; the diffusion study reports mean $\pm$ 95\% CI over 5 seeds. Diffusion generator: Gaussian DDPM with a cosine schedule on quantile-normal-transformed defaulter rows.

The exact base prompt template is shown in Fig.~\ref{fig:prompt}, and the exact condition-specific instructions and hybrid additions in Fig.~\ref{fig:instr}.

\begin{figure}[!t]
\footnotesize
\begin{verbatim}
You are predicting credit-card default (1 = will
default next month, 0 = will not) for UCI 'Default
of Credit Card Clients' data.
PAY_0..PAY_6 = repayment-status codes (>=1 means
months delayed), BILL_AMT* = bills, PAY_AMT* =
payments, LIMIT_BAL = credit limit.
12 labeled examples (features -> label):
- LIMIT_BAL=50000, ..., PAY_AMT6=2000 -> 1
  ... (11 more balanced examples) ...

<CONDITION-SPECIFIC INSTRUCTION>

Clients:
id 0: LIMIT_BAL=20000, ..., PAY_AMT6=0
  ... (up to 20 clients per request) ...

Return ONLY a JSON array, one object per client
in order:
[{"id":0,"label":0,"confidence":0.0}, ...]
confidence = probability that label=1 (default),
0..1. No prose.
\end{verbatim}
\caption{Exact base prompt template (verbatim). Few-shot rows and client rows are abbreviated with ``...''; all other text is sent as shown.}
\label{fig:prompt}
\end{figure}

\begin{figure}[!t]
\scriptsize
\begin{verbatim}
L1 (direct):
  Predict each client's label using general
  repayment-behavior patterns.

L2 (role + features):
  Act as a trained Random Forest classifier. A
  fitted Random Forest ranks these features most
  important (in order): PAY_0, PAY_2, BILL_AMT1,
  PAY_4, PAY_3, LIMIT_BAL, PAY_AMT1, BILL_AMT2.
  Weigh repayment-status (PAY_*) most, then
  bill/payment amounts.

L2a (role only):
  Act as a trained Random Forest classifier. Use
  your own judgment about which features matter most.

L2b (features only):
  When deciding, weigh these features most (in
  order): <top-8 list>. Weigh repayment-status
  (PAY_*) most, then bill/payment amounts.

E1/E3/E5 (feature pruning): same instruction as L1,
  but each row lists ONLY the top-8 features.

E6 (hybrid) preamble adds:
  Each row also includes RF_prob (a trained Random
  Forest's estimated probability of default) and
  RF_pred (its 0/1 decision). Combine RF's opinion
  with the features to decide.
  ... and each row appends:  RF_prob=0.73, RF_pred=1

E7 (ensemble hybrid) appends three classifiers:
  RF_prob=.., RF_pred=.., LR_prob=.., LR_pred=..,
  ADA_prob=.., ADA_pred=..
\end{verbatim}
\caption{Exact condition-specific instructions and hybrid row additions (verbatim). \texttt{<top-8 list>} is the ordered feature list shown in L2.}
\label{fig:instr}
\end{figure}

\medskip
\noindent\textbf{Reproducibility.} All code, data-preparation scripts, per-condition prompts, cached LLM outputs, result tables, and the scripts that produce every figure are available at \url{https://github.com/researchsparkhub/credit_risk_llm} (made public upon publication). Because the LLM responses are cached in the repository, every reported number can be regenerated without new API calls; the dataset is rebuilt from the public UCI source via the included conversion script. Random seeds are fixed (42; diffusion confidence intervals use seeds 0--4).
\end{document}